\documentclass[conference, letterpaper]{IEEEtran}
\IEEEoverridecommandlockouts

\usepackage{cite}
\usepackage{amsmath,amssymb,amsfonts}
\usepackage{algorithmic}
\usepackage{graphicx}
\usepackage{textcomp}
\usepackage{xcolor}
\usepackage[table]{xcolor}
\usepackage{svg}
\usepackage{url}
\usepackage{bm}

\usepackage[hidelinks, colorlinks=true]{hyperref}

\usepackage{xcolor}
\usepackage{pifont}
\usepackage{booktabs} 
\usepackage{multirow}
\usepackage{balance}

\usepackage{geometry}
\usepackage{afterpage}
\afterpage{%
  \newgeometry{top=0.75in, left=0.75in, right=0.75in, bottom=0.75in}
}

\def\BibTeX{{\rm B\kern-.05em{\sc i\kern-.025em b}\kern-.08em
    T\kern-.1667em\lower.7ex\hbox{E}\kern-.125emX}}
\begin{document}

\title{\LARGE \bf LIE: \underline{L}iDAR-only HD Map Construction with \underline{I}ntensity \underline{E}nhancement via Online Knowledge Distillation}


\author{
	\parbox{\textwidth}{%
 		\centering
 		Kanak Mazumder$^{1}$, Fabian B. Flohr $^{1}$
 	}
 	\thanks{$^{1}$Munich University of Applied Sciences, Intelligent Vehicles Lab (IVL), Munich, Germany
 		{\tt\small kanak.mazumder@hm.edu, fabian.flohr@hm.edu}}
}




\maketitle

\begin{abstract}
	Online High-Definition (HD) map construction is a key component of autonomous driving. Recent methods rely on multi-view camera images for cost-effective HD map segmentation, but cameras lack depth information for accurate scene geometry. In contrast, LiDAR provides precise 3D measurements but lacks dense semantic cues. In this work, we propose LIE, LiDAR-only semantic map construction method that employ Knowledge Distillation (KD) to handle the lack of dense semantic and texture cues. Specifically, the teacher branch fuses student LiDAR features and the corresponding 2D intensity map tile to provide dense supervision for segmenting map elements using online distillation scheme. Experimental results show that our method outperforms all single-modality approaches, achieving 8.2\% higher mIoU than the state-of-the-art camera-based model on nuScenes. LIE is robust over long ranges and under challenging weather and lighting, and efficiently adapts to Argoverse2 with only 10\% fine-tuning, surpassing camera-based models trained on the full dataset. Source code will be available \href{https://iv.ee.hm.edu/lie/}{here}.
\end{abstract}


\section{Introduction}
High-definition (HD) maps have become a cornerstone of autonomous driving systems, providing detailed geometric and semantic information about road layouts, lane boundaries, and traffic-related features for downstream tasks such as planning and navigation. The traditional HD map construction pipeline involves building globally consistent point cloud maps using Simultaneous Localization and Mapping (SLAM) from recorded LiDAR data and annotating the map with relevant semantic labels. The construction of such maps entails significant costs and precludes real-time updates~\cite{elghazaly-highdefinition2023,hossain-highdefinition}. Online HD map prediction has reduced the efforts required for annotation and maintenance of the HD map by predicting the map using the on-board sensors. Existing methods for online HD map segmentation can be divided into image-only approaches, LiDAR-only approaches, and fusion-based approaches. Recent advances in online HD map construction have predominantly relied on multi-view camera images, leveraging image-based perception for their dense appearance cues, cost-effectiveness, and the availability of established pretrained backbones. However, camera images lack depth information, which leads to decreased prediction accuracy for map elements as the distance from the ego vehicle increases. Furthermore, generalization across different platforms and datasets remains a significant challenge. Multi-view camera setups differ across autonomous vehicle platforms and standard autonomous driving datasets in terms of the number of cameras, field of view, mounting positions, orientations, and other configuration parameters. Variations in these parameters significantly affect model performance, as state-of-the-art camera-based HD map construction methods tend to overfit to the camera configuration of the training setup due to their learned view transformation modules. As a result, these methods often fail to transfer to new camera configurations, as shown in~\cite{ranganatha-semvecnet2024}.

In contrast to camera-based approaches, LiDAR sensors are better suited for online HD map construction because they capture accurate three dimensional geometry and are naturally represented in bird’s eye view (BEV) coordinates, thereby avoiding overfitting induced by perspective to BEV view transformations. Differences across LiDAR setups, such as mounting position, number of beams, effective distance range, and intensity range, can be mitigated through appropriate normalization and data augmentation strategies.

\begin{figure}[!t]
    \centering
    \includegraphics[width=\columnwidth]{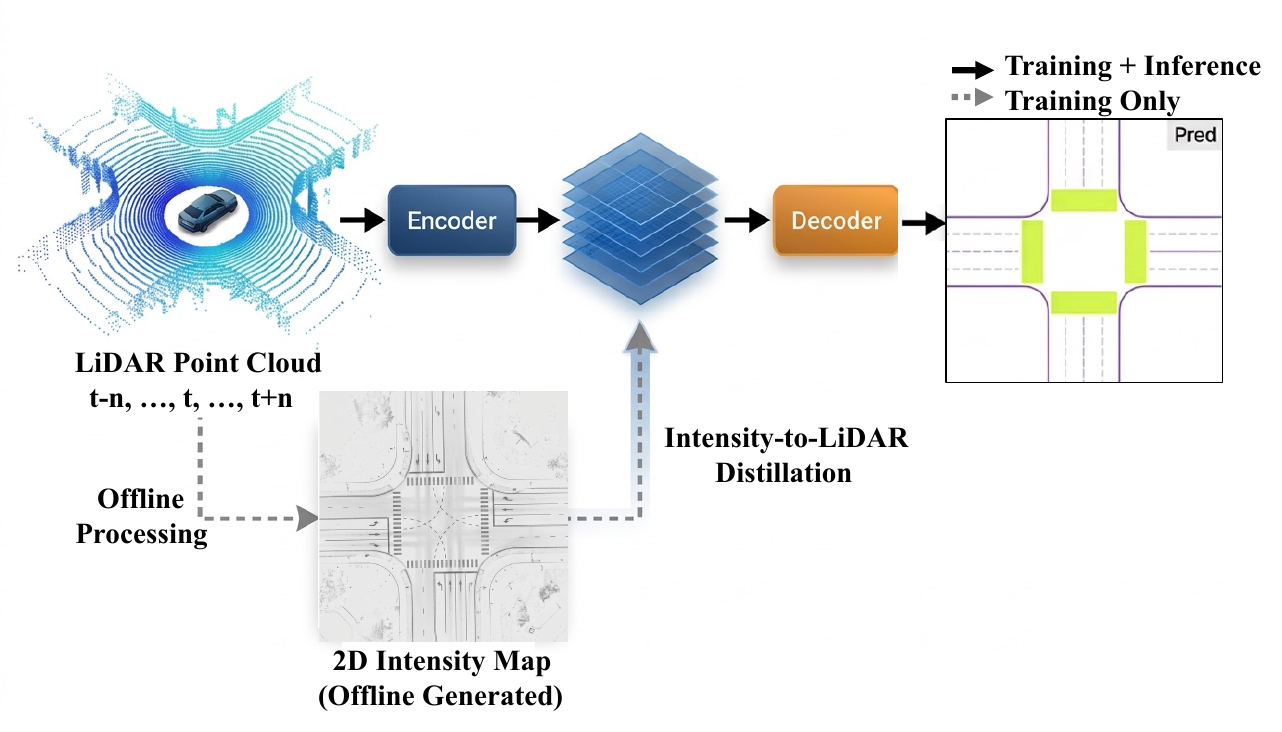}
    \caption{We propose LIE, a LiDAR-only HD map construction framework leveraging Intensity-to-LiDAR distillation using an offline-generated 2D intensity map. Compared with fusion methods, we use dense intensity map only during training, without introducing additional complexity at inference time.
    }
    \label{fig:motivation}
\end{figure}
A key reason for the relative underrepresentation of LiDAR in HD map segmentation lies in the representational gap between sparse, irregular point clouds and the dense semantic features commonly used in segmentation networks~\cite{liu-bevfusion2022}. However, HD map segmentation differs fundamentally from general semantic segmentation. Map elements such as lane dividers, pedestrian crossings, and road boundaries exhibit strong structural regularities, which have been effectively exploited in traditional lane-level road map construction pipelines~\cite{gwon-generation2016,yu-laser2021}. Moreover, many road elements possess distinctive reflectivity patterns that are directly captured by the LiDAR intensity or reflectivity channel, yet this information remains largely underutilized in existing HD map construction methods. Consequently, camera-driven pipelines continue to dominate HD map segmentation benchmarks, despite the fact that LiDAR provides inherently superior geometric and structural information for this task.

In this work, we revisit LiDAR-only HD map construction and focus on online rasterized map segmentation directly from point clouds. To compensate for the absence of dense appearance cues, we introduce a knowledge distillation framework based on 2D LiDAR intensity maps, as illustrated in Fig.~\ref{fig:motivation}. These rasterized intensity projections, available only during training, serve as an intermediate representation that bridges sparse point clouds and dense 2D segmentation paradigms. By providing complementary structural and reflectivity cues without requiring camera inputs, they enhance feature learning while preserving the deployment simplicity of a LiDAR-only system. Consequently, our approach enables LiDAR-based models to acquire fine-grained segmentation capabilities without sacrificing robustness and generalization. Our contributions are summarized as follows:
\begin{itemize}
\item We propose a real-time LiDAR-only framework for online HD map construction that efficiently leverages auxiliary reflectance signals along with geometric range measurements to enhance semantic understanding.
\item We introduce an online Intensity-to-LiDAR distillation strategy to guide feature learning from rasterized intensity maps.
\item We demonstrate strong performance and cross-dataset adaptability, outperforming existing camera and LiDAR-based methods on nuScenes and efficiently transferring to Argoverse2 with minimal fine-tuning.
\end{itemize}


\section{Related Works}

\subsection{Semantic Map Learning.} 
HD maps encode rich semantic and geometric road information and can be represented as rasterized or vectorized maps. From a construction standpoint, existing approaches fall into three main groups: camera-based, LiDAR-based, and fusion-based methods. Camera-based approaches transform perspective-view features to bird’s-eye view (BEV) features via geometric priors~\cite{li-hdmapnet2022,philion-lift2020,harley-simplebev2022} or learned modules~\cite{zhou-crossview2022,li-bevformer2022,liu-petr2022}, but this transformation is ill-posed and struggles to generalize across weather, lighting, and sensor configuration changes. LiDAR-based methods~\cite{li-hdmapnet2022,wang-lidar2map2023a} exploit precise 3D measurements, providing greater robustness under challenging environmental conditions and diverse sensor setups. Fusion methods combine LiDAR and camera features to improve BEV representations~\cite{li-hdmapnet2022,liu-bevfusion2022,kim-broadbev2023}, often sacrificing real-time capability in exchange for higher performance. BroadBEV~\cite{kim-broadbev2023} achieves the highest performance on nuScenes; however, it relies on computationally expensive cross-attention fusion of high resolution feature maps, a nuScenes pre-trained camera backbone, and 3 times more LiDAR sweeps incomparison to baselines.

\subsection{Prior-based Map Learning}
HD map construction focuses on static environmental features, which change infrequently. Leveraging prior information, such as navigation or satellite maps, can improve prediction confidence. P-MapNet~\cite{jiang-pmapnet2024} uses OpenStreetMap road skeletons and HD map priors alongside onboard sensor data to extend perception beyond sensor range. SatforHDMap~\cite{gao-satmap2024} and SatMap~\cite{mazumder-satmap2026}  augment camera features with satellite semantic features to improve HD map accuracy. NMP~\cite{xiong-neural2023} maintains a global sparse map prior, which is updated using fused features from local sensor observations. Despite their advantages, prior-based approaches require accurately aligned maps at inference and introduce additional computational complexity and latency.

\begin{figure*}[!h]
    \centering
    \includegraphics[width=0.95\linewidth]{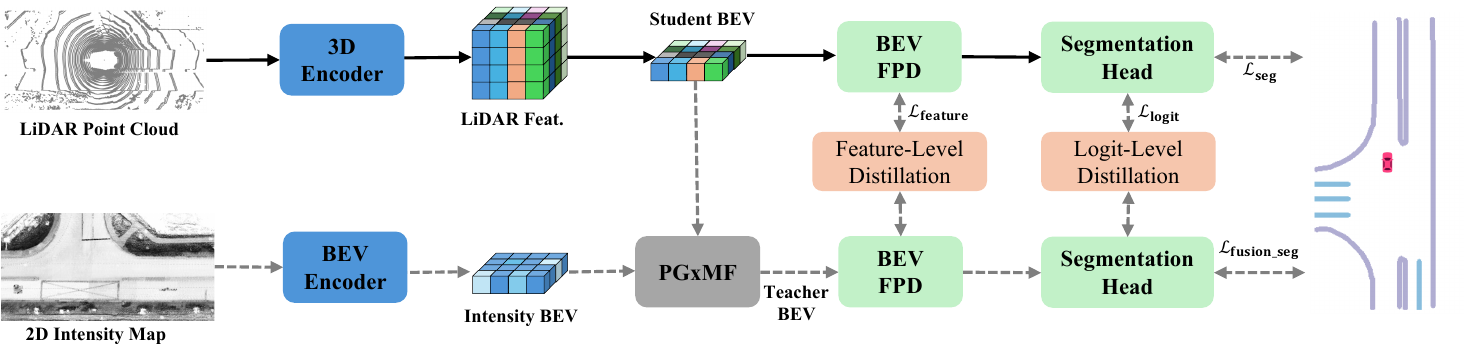} 
    \caption{\textbf{Framework Overview}. LIE uses online feature and logit-level distillation from lidar intensity image to learn lane-related intensity features. Both lidar student branch and lidar-intensity fusion teacher branch use multi-layer BEV decoder. During inference, only the lidar branch is used without any additional overhead.}
    \label{fig:pipeline}
\end{figure*}

\subsection{Knowledge Distillation}
Knowledge distillation (KD) is a training strategy where a smaller model learns to mimic a pre-trained larger model~\cite{Hinton2015DistillingTK}. Originally proposed for model compression, KD has been widely applied in image classification, semantic segmentation, and object detection. In autonomous driving, BEV-based KD methods such as BEVDistill~\cite{chen-bevdistill2022}, DistillBEV~\cite{wang-distillbev2023}, and UniDistill~\cite{zhou-unidistill2023} have shown promising results for 3D object detection. In the context of HD map construction, KD remains underexplored. LiDAR2Map~\cite{wang-lidar2map2023a} distills semantic knowledge from multi-camera features to a LiDAR-based student to reduce the modality gap. Meanwhile, MapDistill~\cite{hao-mapdistill2025} and MapKD~\cite{yan-mapkd2025} train lightweight camera-based map models using multi-modal teachers or teacher–coach–student schemes, leveraging prior maps and fusion knowledge to improve accuracy and reduce latency. 

In contrast to traditional multi-modal distillation, which transfers rich features from multiple modalities to single-modality student model, we employ online \textit{self-distillation} using  offline processed dense LiDAR intensity map to supervise sparse current LiDAR scan. This avoids reliance on full 360° camera coverage during training and allows the student to fully exploit LiDAR’s inherent structural cues rather than mimicking image-based semantics, as in~\cite{wang-lidar2map2023a}, resulting in better performance.


\section{Method}

\subsection{Overview}
In this work, we investigate the potential of the LiDAR modality for real-time HD map segmentation. Despite the current belief that LiDAR lacks semantic cues for accurate semantic segmentation, we argue that a LiDAR-based model can learn relevant semantic information to perform map segmentation. To learn relevant semantic cues, we employ a knowledge distillation scheme from 2D intensity map to sparse LiDAR modality during training. Only the LiDAR branch is deployed in the inference stage for semantic map prediction. Fig.~\ref{fig:pipeline} shows the pipeline of the proposed LIE framework.

\subsection{LiDAR Feature Extractor}
To extract features from LiDAR point cloud, we employ the PointPillars~\cite{lang-pointpillars2019} variant as in~\cite{wang-lidar2map2023a}. In order to address the point cloud sparsity, the point cloud is first voxelized using dynamic voxelization to reduce memory usage. To create the pseudo-image from a point cloud, the voxelized point cloud is converted to a set of pillars. Each point $p$ in the pillar has three-dimensional coordinates, intensity, timestamps, and $K$ features from voxelization represented as $\mathbf{f}_{\text{point}} \in \mathbf{R}^{\text{K}+5}$. Points in each pillar is then processed using a PointNet~\cite{qi-pointnet2017a} and aggregated using a $\mathbf{max}$ operation over all points $N$ in the pillar to get pillar-wise feature $\mathbf{f}_{\text{pillar}}$, where 
\begin{equation}\mathbf{f}_{\text{pillar}} = \max_{\text{point} \in [1, N]} \left( \textbf{ReLU} \left( \textbf{BN} \left( \textbf{Conv} \, 1 \times 1 \, (\mathbf{f}_{\text{point}}) \right) \right) \right).\end{equation}
The encoded pillar features are scattered back to create the pseudo-image of the point cloud. The pseudo-image is further passed through multiple convolution and upsampling blocks to obtain the LiDAR bird-eye-view $\mathbf{F}^\text{BEV}_\text{LiDAR}$.

\subsection{2D Intensity Map Representation and Extraction}
The 2D LiDAR intensity map is a planar grid where each cell encodes the maximum reflectance of LiDAR points within that cell. To focus on the road surface, ground points are first extracted from the 3D point cloud~\cite{shen-jpc2021,lee-patchwork2022}. For scene-level maps, all filtered point clouds in the scene are transformed to a global frame and concatenated to increase density and reduce sparsity. The accumulated points are discretized into Cartesian grid cells, retaining the maximum intensity per cell. To enhance structural features such as lane markings, the resulting map is normalized and refined using local maximum filtering and Gaussian smoothing. A more detailed description can be found in~\cite{yu-laser2021,gwon-generation2016,aldibaja-reliable2020a}. As nuScenes provide city-level intensity map, we extract per-sample intensity map tile based on ego pose and perception range for our experiment. The provided map have a resolution of 0.1m.

\subsection{Intensity Feature Extractor}
We utilize an image encoder to directly extract BEV features, as the intensity map is already represented in bird’s-eye view. Since intensity maps capture road elements of varying sizes, the encoder is designed to leverage multi-scale features. In particular, we adopt Swin Transformer~\cite{liu-swin2021} as the backbone to extract multi-scale feature maps $C_2$, $C_3$, $C_4$, and $C_5$, where $C_i$ corresponds to a spatial resolution of $\frac{H}{2^i} \times \frac{W}{2^i}$. To facilitate effective fusion across these multi-scale features, we employ a generalized FPN~\cite{lin-feature2017} following the approach in~\cite{liu-bevfusion2022}.

\subsection{BEV Decoder}
The BEV features produced by the LiDAR backbone often contain ambiguous noise, which significantly contributes to the relatively lower segmentation performance of LiDAR-based methods. Rather than employing a simple fully connected layer as the segmentation head, we adopt an FPN-based BEV-FPD module to extract and fuse multi-scale BEV features, following the designs in~\cite{wang-lidar2map2023a,kim-broadbev2023}.

\subsection{Online Intensity-to-LiDAR Distillation}
To learn relevant semantic features from LiDAR data, we propose an online Intensity-to-LiDAR knowledge transfer strategy, which guides the LiDAR encoder to distinguish the lane-level road markings. The lane markings, which are easily visible and distinguishable in camera images, are hard to distinguish in sparse LiDAR data. The distillation scheme covers the gap between sparse LiDAR features and dense features relevant for semantic mapping. The distillation scheme has three modules: Feature Fusion Module, Feature-level Distillation, and Logit-level Distillation.

\begin{figure}[!tb]
    \centering
\includegraphics[width=0.9\columnwidth]{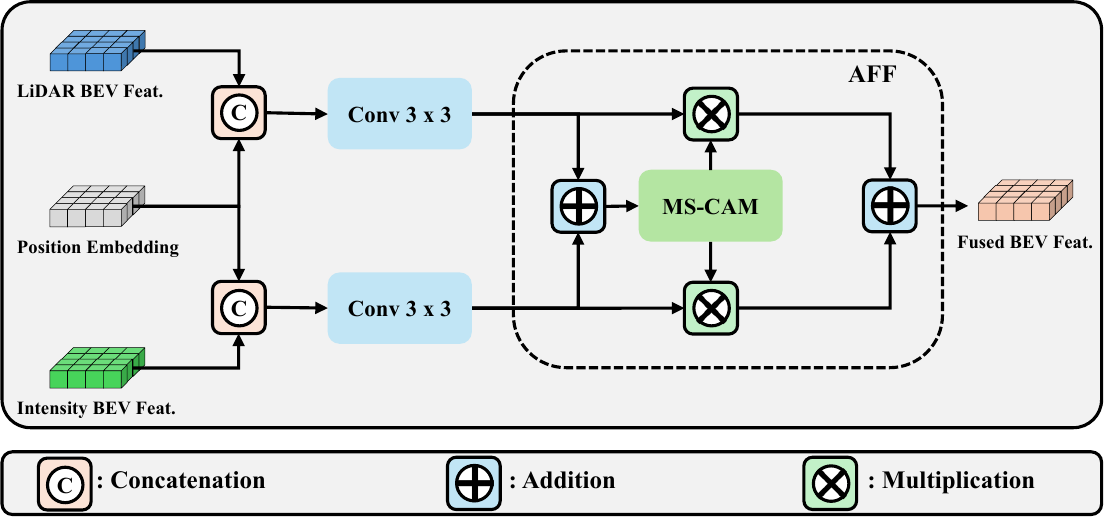} 
    \caption{Illustration of Position-Guided Cross-Modal Fusion (PGxMF) module. In PGxMF, the intensity map features and LiDAR features in BEV space are fused using attentional feature fusion.}
    \label{fig:pgxmf}
\end{figure}

\textbf{Feature Fusion Module.}
To effectively fuse features from the LiDAR and intensity branches, we propose a Position-Guided Cross-Modal Feature (PGxMF) fusion module, inspired by~\cite{wang-lidar2map2023a,dai-attentional2021}. First, a position embedding $\mathbf{F}_\text{PE}$ is computed based on the relative x- and y-coordinates, matching the spatial dimensions of $\mathbf{\tilde F}^\text{BEV}$. This embedding is integrated with the BEV features via concatenation followed by a $3 \times 3$ convolution as follows,
\begin{equation}
\mathbf{\tilde F}^{\text{BEV}}_{\text{PE}} = \textbf{Conv} \, 3 \times 3 \, ([\mathbf{\tilde F}^{\text{BEV}}, \, \mathbf{F}_{\text{PE}}]).
\end{equation}
Next, the position-encoded features $\mathbf{F}^\text{BEV}_{\text{LiDAR\_PE}}$ and $\mathbf{F}^\text{BEV}_{\text{Intensity\_PE}}$ are input to an Attentional Feature Fusion (AFF)~\cite{dai-attentional2021} module, which computes attention weights based on both global and local context, as illustrated in Fig.~\ref{fig:pgxmf}. The resulting fused feature, $\mathbf{F}^\text{BEV}_{\text{Fusion}}$, serves as the BEV representation of the teacher model for distillation.

\textbf{Feature-level Distillation.}
To enable the student branch encoder to learn from dense intensity features, we employ feature distillation using multi-level BEV feature $\{\mathbf{\tilde F}^\text{BEV}_{i}\}^N_{i=1}$ from the BEV decoder, as in~\cite{wang-lidar2map2023a}. To capture long-range dependencies in each modality, a learnable tree filter (LTF)~\cite{song-learnable2019, liang-tree2022} is used. First, minimum spanning trees (MSTs) are built on both low pillar/voxel features and multi-scale BEV features $\{\mathbf{\tilde F}^\text{BEV}_{i}\}^N_{i=1}$ to calculate low-level and high-level affinity matrices as follows, 
\begin{equation}
    \textbf{A} = \text{exp}(-\mathcal{D}(\mathcal{T}(\textbf{F}))),
\end{equation}
where, $\mathcal{T}$ is MST layer,  $\mathcal{D}$ is distance map calculation.
Then, LTF is used as a structure preserving feature transform module to build global affinity matrices from low and high level affinity matrices in a cascaded manner as below,
\begin{equation}
    \tilde{\textbf{M}}^\text{BEV}_{i} = \mathcal{F}(\mathcal{F}(\tilde{\textbf{A}}^\text{BEV}_{i}, {\textbf{A}}^\text{low}), \tilde{\textbf{A}}^\text{BEV}_{i}).
\end{equation}
The feature distillation loss is then calculated from LiDAR affinity matrices and fusion affinity matrices by calculating simple $\mathcal{L}_1$ distance accumulated over all scales as below,
\begin{equation}
    \mathcal{L}_\textbf{feature} = \sum_{i=1}^{N} {\|{\textbf{M}}^\text{BEV}_{\text{Fusion}, i} - {\textbf{M}}^\text{BEV}_{\text{LiDAR}, i} \|}_{1}.
\end{equation}

\textbf{Logit-level Distillation.}
The logit-level distillation module introduces an auxiliary training objective in which the student model is guided to replicate the teacher model’s semantic predictions. Rather than focusing solely on the class with the highest probability, the student learn a soft label by matching the whole distribution. As in~\cite{wang-lidar2map2023a,jaritz-xmuda2020}, we employ KL divergence to calculate the similarity loss between the prediction distributions of the teacher and the student, as defined below,
\begin{equation}
    \mathcal{L}_\textbf{logit} = \textbf{D}_\text{KL} (\mathbf{P}^\text{BEV}_{\text{Fusion}} \| \mathbf{P}^\text{BEV}_{\text{LiDAR}}).
\end{equation}

\subsection{Training and Inference}
\textbf{Overall Training Loss.} 
This work addresses the rasterized semantic HD map construction task by optimizing the model using segmentation loss. To enable effective knowledge propagation from teacher model to student model, we combine teacher segmentation loss, student segmentation loss with the distillation losses mentioned above, including multi-level feature distillation loss and logit distillation loss, following~\cite{wang-lidar2map2023a}. The complete training loss can be expressed as:
\begin{equation}
    \mathcal{L} = \mathcal{L}_{\textbf{fusion\_seg}} + \mathcal{L}_\textbf{seg} + \alpha * \mathcal{L}_\textbf{feature} + \beta * \mathcal{L}_\textbf{logit}
    \label{}
\end{equation}
The segmentation loss $\mathcal{L}_{\textbf{fusion\_seg}}$ and $\mathcal{L}_\textbf{seg}$ are calculated following~\cite{wang-lidar2map2023a}, which consist of cross-entropy loss and Lov\'asz-Softmax loss~\cite{berman-lovaszsoftmax2018}.

\setlength{\tabcolsep}{5pt} 
\begin{table*}[!h]
\caption{Performance comparison on the original \textit{val} set of nuScenes with the 60m$\times$ 30m map segmentation setting. ``C'' and ``L'' denote surround-view camera and LiDAR inputs, respectively;``SD'' and``HD'' represent Standard-Definition and High-Definition maps; and ``L\textdagger'' indicates the 2D intensity map from LiDAR. ``*'' marks UniFusion~\cite{qin-unifusion2023} reported results.}
\centering
\begin{tabular}{l c c c c c c c}
\hline
\rowcolor{gray!10}
\textbf{Method} & \textbf{Modality} & \textbf{Teacher Modality} & \textbf{Backbone} & \textbf{Divider} & \textbf{Ped Crossing} & \textbf{Boundary} & \textbf{mIoU} \\
\hline
HDMapNet-Camera~\cite{li-hdmapnet2022} & C & - & EfficientNet-B0 & 40.5 & 19.7 & 40.5 & 33.6 \\
DiffMap~\cite{jia-diffmap2024} & C & - & EfficientNet-B0  & 42.1 & 23.9 & 42.2 & 36.1 \\
MapKD~\cite{yan-mapkd2025} & C & C+L+SD+HD & EfficientNet-B0 & 25.4 & 44.4 & 43.7 & 37.8 \\
BEVSegFormer*~\cite{peng-bevsegformer2023} & C & - & ResNet-101 & 51.1 & 32.6 & 50.0 & 44.6 \\ 
BEVFormer*~\cite{li-bevformer2022} & C & - & ResNet-50 & 53.0 & 36.6 & 54.1 & 47.9 \\
BEVFormer + NMP~\cite{xiong-neural2023} & C & - & ResNet50 & 55.0 & 34.1 & 56.5 & 48.5 \\
BEVerse*~\cite{zhang-beverse2022} & C & - & Swin-T & 56.1 & \textbf{44.9} & 58.7 & 53.2 \\
UniFusion*~\cite{qin-unifusion2023} & C & - & Swin-T & \textbf{58.6} & 43.4 & \textbf{59.0} & \textbf{53.6} \\
\hline
HDMapNet-Fusion~\cite{li-hdmapnet2022} & C+L & - & EfficientNet-B0 \& PointPillars & 46.1 & 31.4 & 56.0 & 44.5 \\
DiffMap~\cite{jia-diffmap2024} & C+L & - & EfficientNet-B0 \& PointPillars & 54.3 & 34.4 & 60.7 & 49.8 \\
P-MapNet~\cite{jiang-pmapnet2024} & C+L+SD+HD & - & EfficientNet-B0 \& PointPillars & 54.2 & 41.3 & 63.7 & 53.1 \\
SatforHD~\cite{gao-satmap2024} & C+Satellite & - & EfficientNet-B0 \& ResNetUNet & 54.9 & 53.4 & 52.9 & 53.7 \\
LiDAR2Map~\cite{wang-lidar2map2023a} & C+L & - & Swin-T \& PointPillars & 60.8 & 47.2 & 66.3 & 58.1 \\
BroadBEV~\cite{kim-broadbev2023} & C+L & - & Swin-T \& VoxelNet & \textbf{68.8} & \textbf{51.2} & \textbf{71.9} & \textbf{64.0} \\
\hline
HDMapNet-LiDAR~\cite{li-hdmapnet2022} & L & - & PointPillars & 26.7 & 17.3 & 44.6 & 29.5 \\
LiDAR2Map~\cite{wang-lidar2map2023a} & L & C & PointPillars & 60.4 & 45.5 & \textbf{66.4} & 57.4 \\

\rowcolor{gray!20}
LIE (ours) & L & L\textdagger & PointPillars & \textbf{61.0} & \textbf{46.6} & 66.3 & \textbf{58.0} \\
\hline
\end{tabular}
\label{tab:mapseg_results}
\end{table*}

\textbf{Inference.} During inference, the pipeline uses only the LiDAR branch, avoiding any extra computational overhead from the distillation loss or the intensity–LiDAR fusion branch. The LiDAR branch, trained to capture both geometric information and dense intensity features, effectively encodes the static map elements.

\section{Experiments}

\textbf{Dataset.} We conduct evaluations on the widely recognized and challenging nuScenes~\cite{caesar-nuscenes2020} dataset, a large-scale autonomous driving benchmark that includes annotated HD semantic maps with precise localization, collected in Boston and Singapore. The dataset is composed of three subsets: 700 scenes for training, 150 scenes for validation, and 150 scenes for testing, where each scene has a duration of 20 seconds. The camera data is captured by six cameras providing a 360-degree field of view, each operating at 12 Hz, alongside LiDAR data from a 32-beam LiDAR operating at 20 Hz.
As recent works~\cite{yuan-streammapnet2024, lilja-localization2024a} have identified substantial geographical overlap between the original training and validation splits of nuScenes, this setup risks location memorization rather than true generalization in HD map construction. Nevertheless, since most semantic HD map methods report results only on the original split, we provide evaluations on the official split and compare our results with those reported on the geospatially non-overlapping split introduced in~\cite{yuan-streammapnet2024}.
We further experiment on Argoverse2~\cite{wilson-argoverse22023}, which contains 1000 driving scenes from six different US cities, each with 15 seconds duration. The data is collected with 7 high resolution camera operating at 20 Hz and 2$\times$32 beam (effectively 64-beam) LiDAR operating at 10 Hz.

\textbf{Metric.} The performance of our semantic map construction method is assessed using Mean Intersection over Union (mIoU), following the protocol in~\cite{li-hdmapnet2022}. Evaluations are conducted over both standard and extended perception ranges around the ego vehicle. Consistent with HDMapNet~\cite{li-hdmapnet2022}, we focus on three static map elements: lane dividers, pedestrian crossings, and road boundaries. To compare vectorized map result, Average Precision (AP) is adopted as the evaluation metric, as proposed in~\cite{li-hdmapnet2022,liu-vectormapnet2022}. The AP caculations are calculated under distance threshold of 0.5m, 1.0m, 1.5m.

\textbf{Model and Training Details.} We use Swin-T~\cite{liu-swin2021}, pretrained on ImageNet~\cite{deng-imagenet2009}, as our backbone for intensity feature extraction. We train the whole network end-to-end using the Adam optimizer~\cite{kingma-adam2017} for 30 epochs with weight decay of $1e^{-7}$ on 4 NVIDIA A40 GPUs. The learning rate is $2e^{-3}$ with StepLR, which decreases the learning rate by a factor of 10 at 20 epochs. The training setup follows \cite{wang-lidar2map2023a} for better comparison. The BEV resolution for standard perception range of 60m$\times$30m is 0.15m  and the resolution of 0.3m is used for extended range of 120m$\times$60m. The $\alpha$ and $\beta$ parameters in our experiments are set to 0.4 and 1.5 respectively.

\begin{figure*}[!h]
    \centering
    \includegraphics[width=0.9\linewidth]{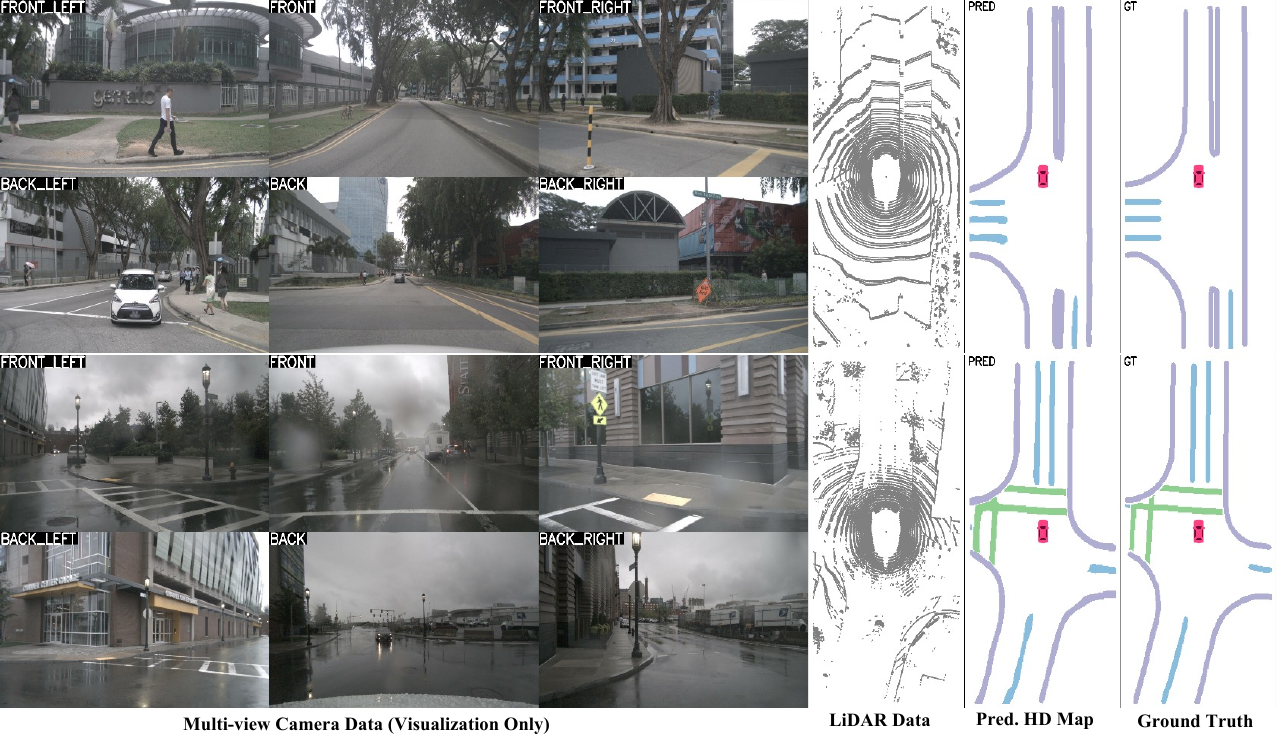} 
    \caption{Visualization of LIE on the nuScenes original \textit{val} set under cloudy and rainy conditions within a $60\text{m} \times 30\text{m}$ perception range. The left column shows the six surrounding camera views, provided solely for visualization. The middle column displays the BEV density image of input LiDAR pointcloud used for inference, while the right column presents the predicted semantic map alongside the corresponding ground truth.}
    \label{fig:result}
\end{figure*}

\textbf{Comparison with the State-of-the-Art.} We evaluate our approach against leading camera-based, fusion-based, and LiDAR-based methods, as well as existing knowledge distillation and diffusion approaches for semantic HD map construction. Our method outperforms all camera-based HD map segmentation models, achieving 8.2\% mIoU improvement over UniFusion~\cite{qin-unifusion2023} (see Table~\ref{tab:mapseg_results}). Compared to most fusion methods, performance is comparable or better except for BroadBEV~\cite{kim-broadbev2023}, which is computationally demanding and has high latency (6.3 FPS on NVIDIA A100). Furthermore, its excessive memory requirements prevent single-sample inference on consumer-grade GPUs, whereas our method runs at over 36 FPS on a single NVIDIA RTX 4090. LIE surpasses LiDAR2Map~\cite{wang-lidar2map2023a} by 1\% mIoU using the same student backbone without multi-view camera distillation, and achieves performance just 0.17\% mIoU below LiDAR2Map’s camera–LiDAR fusion model while running at twice the speed. Our teacher model is lightweight, achieving 29 FPS in evaluation mode; this is for complexity analysis only, as it uses offline scene or city level intensity maps. It is suitable when prior intensity maps are available or high-resolution LiDAR enables per-frame intensity generation.

Notably, LIE performs particularly well on lane dividers and pedestrian crossings, which have distinctive LiDAR reflectivity patterns, highlighting the benefit of intensity-based supervision. Qualitative results in representative driving scenarios are shown in Fig.~\ref{fig:result}.

\textbf{Long Range Experiment.} 
We conduct a performance comparison with existing long-range HD map construction methods. We compare our method with~\cite{li-hdmapnet2022} and P-MapNet~\cite{jiang-pmapnet2024} in the 120m$\times$60m perception range, as shown in Table~\ref{tab:mapseg_results_120x60}. 
Our method outperforms HDMapNet Camera and Camera-LiDAR fusion methods. In comparison with P-MapNet, which is specially designed for long-range perception, our method outperforms their camera-based models with SD and HD prior maps.
\setlength{\tabcolsep}{5pt} 
\begin{table}[!h]
\caption{Comparison of map segmentation methods on original nuScenes \textit{val} split with 120m$\times$60m perception range. ``*" marks P-MapNet~\cite{jiang-pmapnet2024} reported results.}
\centering
\begin{tabular}{lcccccc}
\hline
\rowcolor{gray!10}
\textbf{Method} & \textbf{Mod.} & \textbf{Div.} & \textbf{Ped.} & \textbf{Bound.} & \textbf{mIoU} \\
\hline
HDMapNet*~\cite{li-hdmapnet2022} & C & 39.2 & 23.0  & 39.1 & 33.8 \\
P-MapNet*~\cite{jiang-pmapnet2024} & C+SD & 44.8 & 30.6 & 45.6  & 40.3  \\
P-MapNet*~\cite{jiang-pmapnet2024} & C+SD+HD & 45.5 & 30.9 & 46.2  & 40.9  \\
HDMapNet*~\cite{li-hdmapnet2022} & C+L & 53.6 & 37.8  & 57.1  & 49.5  \\
\rowcolor{gray!30}
LIE (ours) & L & $\textbf{59.2}$ & $\textbf{45.4}$ & $\textbf{62.4}$ & $\textbf{55.7}$ \\
\hline
\end{tabular}
\label{tab:mapseg_results_120x60}
\end{table}

\textbf{Comparison Under Different Weather and Lighting Conditions.}
In this experiment, we evaluate our method under adverse weather and lighting conditions, as reported in Table~\ref{tab:mapseg_results_weather}. The weather-based validation splits are constructed according to the scene-level descriptions provided in nuScenes. Compared with \cite{wang-lidar2map2023a}, our method exhibits slightly lower performance on these weather-specific subsets. 

\begin{figure}[!t]
    \centering
    \includegraphics[width=\columnwidth]{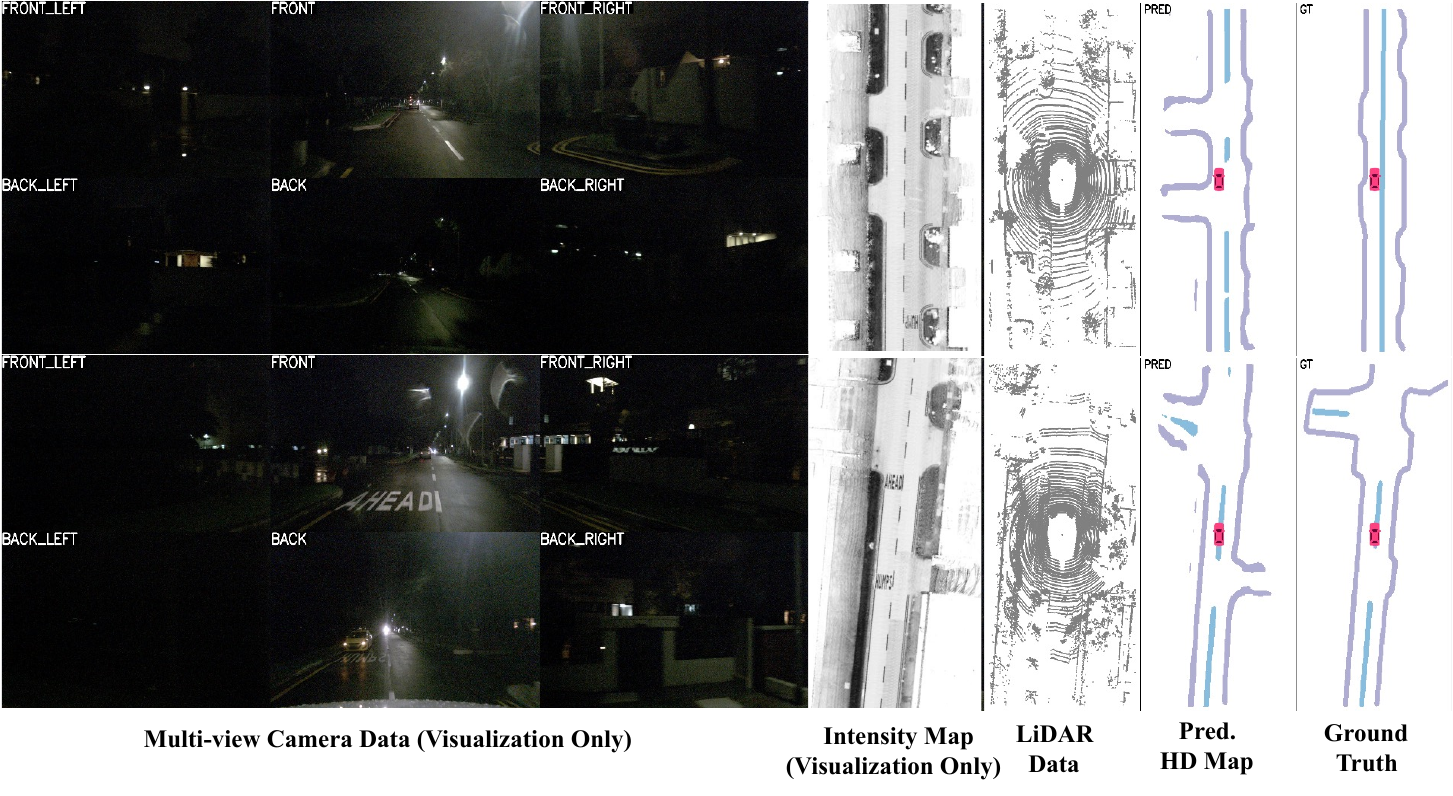} 
    \caption{Qualitative comparison between prediction and ground truth HD map in weather-specific validation split. The prediction tends to follow the geometric structure as visualized in 2D intensity map and LiDAR data, which does not match GT annotation, resulting quantitative performance drop.}
    \label{fig:weather_split}
\end{figure}

A detailed examination of the validation data reveals that all scenes in the night split and several scenes in the rainy split originate from the same residential neighborhood. These areas are characterized by the frequent presence of driveways and private access roads. As visualized in Fig~\ref{fig:weather_split}, driveways and private access roads are labeled as closed road boundaries, whereas access roads leading to underground parking facilities are annotated as road in the ground truth. This discrepancy introduces an inherent inconsistency in the supervision signal during training. In particular, topologically and geometrically analogous regions may correspond to distinct semantic classes, potentially impairing the model’s ability to learn a consistent mapping between geometry and boundary semantics. Our model tends to generate geometrically coherent and topologically plausible road boundary predictions in such scenarios; however, these predictions do not align with the dataset-specific annotation protocol. As a result, the quantitative performance on the corresponding weather splits is adversely affected.

\setlength{\tabcolsep}{5pt} 
\begin{table}[!h]
\caption{Comparison of map segmentation methods on original nuScenes \textit{val} under different weather and lighting conditions in 60m$\times$30m perception range. ``*" marks LiDAR2Map~\cite{wang-lidar2map2023a} reported result.}
\centering
\begin{tabular}{lcccccc}
\hline
\rowcolor{gray!10}
\textbf{Method} & \textbf{Modality} & \textbf{Rainy} & \textbf{Night} & \textbf{All} \\
\hline
HDMapNet-Fusion*~\cite{li-hdmapnet2022} & C + L & 38.7 & 39.3 & 44.5 \\
BEVerse*~\cite{zhang-beverse2022} & C & 48.8 & 44.4 & 51.7 \\
LiDAR2Map~\cite{wang-lidar2map2023a} & L & $\textbf{49.6}$ & $\textbf{49.2}$ & 57.4 \\
\rowcolor{gray!30}
LIE (ours) & L & 49.1 & 48.6 & $\textbf{58.0}$\\
\hline
\end{tabular}
\label{tab:mapseg_results_weather}
\end{table}

\textbf{Vectorization Results.} We further evaluate vectorization performance by applying post-processing to generate vectorized HD map representations, following~\cite{li-hdmapnet2022, jiang-pmapnet2024}. In addition to the post-processed vectorized results from HDMapNet~\cite{li-hdmapnet2022} and P-MapNet~\cite{jiang-pmapnet2024}, we compare against baseline vectorized map prediction methods, including VectorMapNet~\cite{liu-vectormapnet2022} and MapTR~\cite{liao-maptr2023}, as shown in Table~\ref{tab:mapvector_results_60x30}. Although our model is trained for rasterized segmentation, it performs competitively with dedicated vectorized approaches. Under the geospatially non-overlapping split, performance decreases for all methods; however, our model remains competitive with SOTA single-frame deterministic prediction model MapTRv2~\cite{liao-maptrv22024} and maintains consistent AP across all classes.

\setlength{\tabcolsep}{4pt} 
\begin{table}[!h]
\caption{Comparison with various map vectorization methods on nuScenes original \textit{val} split (Org.) and geospatially non-overlapping split (New) at 60m$\times$30m perception range. The AP thresholds are {0.5m, 1.0m, 1.5m}. ``*" denotes result from ~\cite{yuan-streammapnet2024}.
}
\centering
\begin{tabular}{lcccccc}
\hline
\rowcolor{gray!10}
\textbf{Method} & \textbf{Mod.} & \textbf{Split} & \textbf{Div.} & \textbf{Ped.} & \textbf{Bound.} & \textbf{mAP}\\
\hline
HDMapNet~\cite{li-hdmapnet2022} & C & Org. & 21.7 & 14.4  & 33.0 & 23.0 \\
HDMapNet~\cite{li-hdmapnet2022} & C+L & Org. & 29.6 & 16.3  & 46.7  & 31.0 \\
VectorMapNet~\cite{liu-vectormapnet2022} & C & Org. & 47.3 & 36.1 & 39.3 & 40.9 \\
VectorMapNet~\cite{liu-vectormapnet2022} & C+L & Org. & 50.5 & 37.6 & 47.5 & 45.2 \\
MapTR~\cite{liao-maptr2023} & C & Org. & 51.5 & 46.3 & 53.1 & 50.3 \\
\rowcolor{gray!30}
LIE (ours) & L & Org. & \textbf{57.8} & \textbf{48.0} & \textbf{53.9} & \textbf{53.2} \\
\hline
VectorMapNet*~\cite{liu-vectormapnet2022} & C & New & 17.0 & 15.8 & 21.2 & 18.0 \\
MapTR*~\cite{liao-maptr2023} & C & New & 20.7 & 6.4 & 35.5 & 20.9 \\
MapTRv2~\cite{liao-maptrv22024} & C & New & \textbf{28.7} & 16.2 & \textbf{44.8} & \textbf{29.9} \\
\rowcolor{gray!30}
LIE (ours) & L & New & 27.1 & \textbf{23.0} & 34.3 & 28.2 \\
\hline
\end{tabular}
\label{tab:mapvector_results_60x30}
\end{table}

\textbf{Cross-Dataset Experiments.}
To assess the generalization capability of our LIE framework, we trained the model exclusively on the nuScenes dataset and evaluated it on the Argoverse2~\cite{wilson-argoverse22023} validation set. Despite the domain shift and differences in sensor configurations (e.g., different mounting heights and beam numbers), LIE maintains reasonable performance, demonstrating potential cross-dataset transferability. The low performance on pedestrian crossing segmentation comes from the  domain gap in road network distribution, as also observed in~\cite{ranganatha-semvecnet2024}. Additionally, we fine-tuned our student model on just 10\% of Argoverse2 training data, which outperforms HDMapNet~\cite{li-hdmapnet2022} and P-MapNet~\cite{jiang-pmapnet2024} trained on Argoverse2 full dataset, as shown in Table~\ref{tab:av2_results_60x30} , indicating efficient adaptation with limited data.

\setlength{\tabcolsep}{5pt} 
\begin{table}[!h]
\caption{Cross-dataset evaluation on Argoverse2 validation dataset in 60m$\times$30m perception range."FT" denotes finetuning and "Ep." denotes number of training epochs. The best results are highlighted in bold and second best results are underlined.}
\centering
\begin{tabular}{lcccccc}
\hline
\rowcolor{gray!10}
\textbf{Method} & \textbf{Ep.} & \textbf{Mod.} & \textbf{Div.} & \textbf{Ped.} & \textbf{Bound.} & \textbf{mIoU} \\
\hline
HDMapNet~\cite{li-hdmapnet2022} & 6 & C & 53.0 & 27.9  & 44.5 & 41.8 \\
P-MapNet~\cite{jiang-pmapnet2024} & 6 & C+SD & 52.9 & 29.7 & 46.8  & 43.1  \\
P-MapNet~\cite{jiang-pmapnet2024} & 6 & C+SD+HD & 53.5 & \underline{30.1} & 47.3  & 43.6  \\
\rowcolor{gray!30}
LIE (no FT) & - & L & 24.0 & 11.4 & 34.1 & 23.2 \\
\rowcolor{gray!30}
LIE (FT 10\%) & 1 & L & \underline{53.6} & 29.4 & \underline{56.4} & \underline{46.5} \\
\rowcolor{gray!30}
LIE (FT 10\%) & 6 & L & \textbf{58.6} & \textbf{33.6} & \textbf{60.2} & \textbf{50.8} \\
\hline
\end{tabular}
\label{tab:av2_results_60x30}
\end{table}

\textbf{Ablation Studies.} We evaluate the effectiveness of online Intensity-to-LiDAR distillation by analyzing each component of our framework. Our baseline, based on~\cite{wang-lidar2map2023a} with HDMapNet and a 6-layer BEV-FPD decoder, achieves 56.5 mIoU (Table~\ref{tab:ablation}). Logit-level distillation from the Intensity-LiDAR fusion branch increases performance to 56.8 mIoU, while encoder-only feature distillation adds 0.5 mIoU. Multi-level decoder feature distillation yields a 1.1 mIoU gain. Distillation at both encoder and decoder levels reduces performance, likely due to feature instability at early layers. Our Pose-Guided Cross-Model Feature Fusion module, combined with decoder and logit-level distillation, achieves the best performance of 58.0 mIoU.
\begin{table}[!h]
\caption{Ablation study on the effectiveness of different modules of online Intensity-to-LiDAR distillation under various settings at 60m$\times$30m perception range. ``LD", ``ED", and ``DD" correspond to logit-level, encoder feature, and decoder feature distillation, respectively.}
\centering
\begin{tabular}{l|cccc|c}
\hline
\rowcolor{gray!10}
\textbf{Baseline} & \textbf{LD} & \textbf{ED} & \textbf{DD} & \textbf{PGxMF} & \textbf{mIoU} \\
\hline
\checkmark & & & & & 56.5\\
\checkmark & \checkmark & & & & 56.8\\
\checkmark & \checkmark & \checkmark & & & 57.3 \\
\checkmark & \checkmark & & \checkmark & & 57.9 \\
\checkmark & \checkmark & \checkmark & \checkmark & & 57.7 \\
\rowcolor{gray!30}
\checkmark & \checkmark & & \checkmark & \checkmark & 58.0 \\
\hline
\end{tabular}   
\label{tab:ablation}
\end{table}



\section{Conclusion and Future Work}
We presented LIE, a LiDAR-only framework for online HD map construction that leverages online Intensity-to-LiDAR distillation to enhance segmentation from sparse point clouds. By using rasterized intensity maps as training-time guidance, LIE captures fine-grained structural cues without camera inputs, achieving superior performance over existing camera-based and LiDAR-only methods on nuScenes while maintaining real-time inference. Cross-dataset evaluation on Argoverse2 further demonstrates efficient adaptation with limited fine-tuning, highlighting the transferability of LiDAR based approach.

Our current framework does not incorporate sensor-specific data augmentation or multi-dataset training, which may limit generalization. Future work will explore multi-dataset training, temporal fusion, and a real-time camera–LiDAR fusion framework that fully exploits each modality under diverse sensor configurations.


\input{bib/bib-long.def}
\bibliographystyle{bibtex/IEEEtran}
\bibliography{bib/mainBib}

\end{document}